\documentclass[11pt,a4paper]{article}
\usepackage[hyperref]{naaclhlt2018}
\usepackage{times}
\usepackage{latexsym}
\usepackage{graphicx}
\usepackage{subfigure}
\usepackage{url}
\usepackage[utf8]{inputenc}
\usepackage[T2A,LAE,T1]{fontenc}
\usepackage[arabic,english]{babel}
\aclfinalcopy 
\usepackage{color}

\title{ClaimRank: Detecting Check-Worthy Claims in Arabic and English}
\author{%
Israa Jaradat$^1$, 
Pepa Gencheva$^2$  \\ 
\textbf{Alberto Barr\'on-Cede\~no}$^1$, 
\textbf{Llu\'is M\`arquez}$^3$\thanks{\ \ Work conducted while this author was at QCRI.} \and 
\textbf{Preslav Nakov}$^1$ \\
$^1$ Qatar Computing Research Institute, HBKU, Qatar \\ 
$^2$ Sofia University ``St. Kliment Ohridski'', Bulgaria\\ 
$^3$ Amazon, Barcelona, Spain\\
\{ijaradat, albarron, pnakov\}@hbku.edu.qa \hspace{3mm} pepa.k.gencheva@gmail.com \hspace{3mm} lluismv@amazon.com
}

\date{}

\begin{document}

\maketitle
\begin{abstract}
We present ClaimRank, an online system for detecting check-worthy claims. While originally trained on political debates, the system can work for any kind of text, e.g., interviews or regular news articles. Its aim is to facilitate manual fact-checking efforts by prioritizing the claims that fact-checkers should consider first. ClaimRank supports both Arabic and English, it is trained on actual annotations from nine reputable fact-checking organizations (PolitiFact, FactCheck, ABC, CNN, NPR, NYT, Chicago Tribune, The Guardian, and Washington Post), and thus it can mimic the 
claim selection strategies for each and any of them, as well as for the union of them all.
\end{abstract}

\section{Introduction}
\label{sec:intro}

The proliferation of fake news demands
the attention of both investigative journalists and scientists. The need for automated fact-checking systems rises from the fact that manual fact-checking is both effort- and time-consuming. The first step towards building an automated fact-checking system is to identify the claims that are worth fact-checking. 

We introduce ClaimRank, an automatic system to detect check-worthy claims in a given text. ClaimRank is multilingual and at the moment it is available for both English and Arabic. To the best of our knowledge, it is the only such system available for Arabic. 
ClaimRank is trained on actual annotations from nine reputable fact-checking organizations (PolitiFact, FactCheck, ABC, CNN, NPR, NYT, Chicago Tribune, The Guardian, and Washington Post), and thus it can be used to predict the claims by each of the individual sources, as well as their union. This is the only system we are aware of that offers such a capability.

\section{Related Work}

ClaimBuster is the first work to target check-worthiness~\cite{Hassan:15}. It is trained on data annotated by students, professors, and journalists,
and uses features such as sentiment, TF.IDF-weighted words, part-of-speech tags, and named entities. 
In contrast, (\emph{i})~we have much richer features, (\emph{ii})~we support English and Arabic, (\emph{iii})~we learn from choices made by nine reputable fact-checking organizations, and (\emph{iv})~we can mimic the selection strategy of each of them.



In our previous work, we focused on debates from the US 2016 Presidential Campaign and we used pre-existing annotations from online fact-checking reports by professional journalists~\cite{gencheva-EtAl:2017:RANLP}.
Here we use roughly the same features, with some differences (see below).
However, (\emph{i})~we train on more debates (seven instead of four for English, and also Arabic translations for two debates), (\emph{ii})~we add support for Arabic, and (\emph{iii})~we deploy a working system.

\citet{Patwari:17} focused on the 2016 US Election campaign as well and independently obtained their data in a similar way. 
However, they used less features, they did not mimic any specific website, nor did they deploy a working system.


\section{System Overview}
\label{sec:system}

The run-time model is trained on seven English political debates and on the Arabic translations of two of the English debates. For evaluation purposes, we need to reserve some data for testing, and thus the model is trained on five English debates, and tested on the other two (either original English or their Arabic translations). In both cases, the data is first preprocessed and passed to the feature extraction module.
The feature vectors are then fed to the model to generate predictions.

\begin{figure}
\footnotesize
\centering
\scalebox{0.5}{\includegraphics[keepaspectratio]{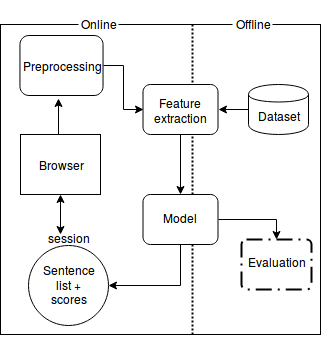}}
\caption{System architecture.}
\label{fig:pipeline}
\end{figure}
\subsection{General Architecture}

Figure~\ref{fig:pipeline} illustrates our general architecture. ClaimRank is accessible via a Web browser. When a user submits a text, the server handles the request by first detecting the language of the text using Python's {\tt langdetect}. Then, the text is split into sentences using NLTK for English and a custom splitter for Arabic. An instance of the sentence list is stored in a session after being JSON-fied. After that, features are extracted for each sentence and fed into the model, which in turn generates the check-worthiness score for each sentence. Scores are displayed in the client next to each sentence, along with their corresponding color codes. Scores are also stored in the session object along with the sentence list as parallel arrays. In case the user wants the sentences sorted by their scores, or wants to mimic one of the annotation sources strategy in sentence selection, the server gets the text from the session, and re-scores/orders it and sends it back to the client.

\subsection{Features}

Here we do not propose new features, but rather reuse features that have been previously shown to work well for check-worthiness~\cite{Hassan:15,gencheva-EtAl:2017:RANLP}. 

From~\cite{Hassan:15}, we include TF.IDF-weighted bag of words, 
part-of-speech tags,
named entities
as recognized by Alchemy API, 
sentiment scores, 
and sentence length (in tokens).

From~\cite{gencheva-EtAl:2017:RANLP}, we adopt lexicon features, e.g., for bias~\cite{Recasens+al:13a}, for sentiment~\cite{Liu:2005:OOA:1060745.1060797}, for assertiveness~\cite{hooper1974assertive}, and also for subjectivity.

\noindent We further use structural features, e.g., for location of the sentence within the debate/intervention, LDA topics~\cite{blei2003latent}, word embeddings~\cite{mikolov2013distributed}, and discourse relations with respect to the neighboring sentences~\cite{jotycodra}.
More detail about the features can be found in the corresponding paper.

\subsection{Model}

In order to rank the English claims,
we re-use the model from \cite{gencheva-EtAl:2017:RANLP}.  In particular, we use a neural network with two hidden layers. We provide the features, which give information not only about the claim but also about its context, as an input to the network. The input layer is followed by the first hidden layer, which is composed of two hundred ReLU neurons \cite{pmlr-v15-glorot11a}. The second hidden layer contains fifty neurons with the same ReLU activation function. Finally, there is a sigmoid unit, which classifies the sentence as check-worthy or not. 

Apart from the class prediction, we also need to rank the claims based on the likelihood of their check-worthiness. For this, we use the probability that the model assigns to a claim to belong to the positive class. We train the model for 100 iterations using Stochastic Gradient Descent \cite{lecun1998gradient}.

\subsection{Adaptation to Arabic}

To handle Arabic along with English, we integrated some new tools. First, we had to add a language detector in order to use the appropriate sentence tokenizer for each language. For English, NLTK's \cite{bird2004nltk} {\tt sent\_tokenize} handles splitting the text into sentences. However, for Arabic it can only split text based on the presence of the period (.) character. This is because other sentence endings ---such as question marks--- are different characters (e.g.,~the Arabic question mark is `\foreignlanguage{arabic}{؟}', and not `?'). Hence, we used our custom regular expressions to split the Arabic text into sentences.

\begin{figure*}	
\footnotesize
\centering
\includegraphics[width=\textwidth,keepaspectratio]{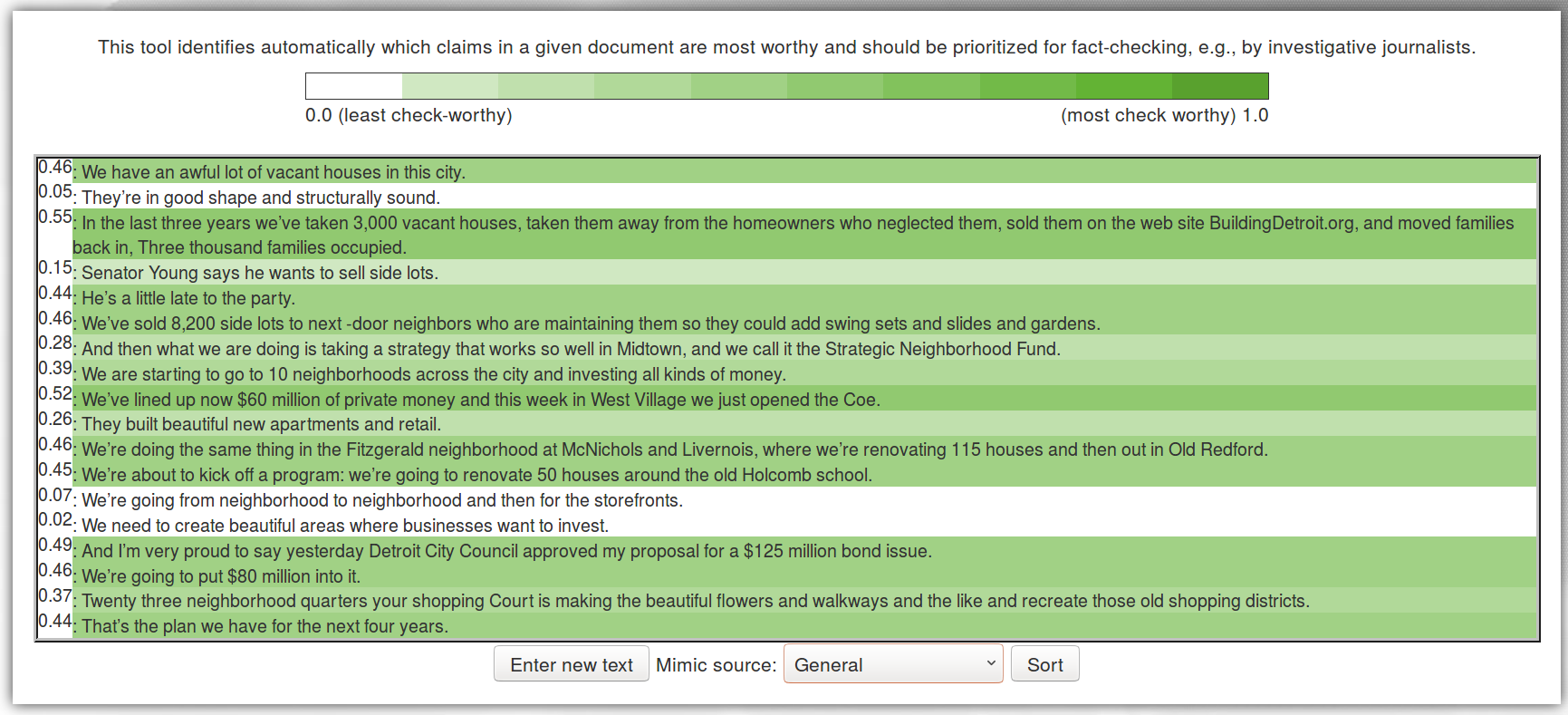}
\caption{\label{fig:demoEnglish}Screenshot of ClaimRank's output for an English presidential debate, in natural order.}
\end{figure*}

Next comes tokenization. 
For English, we used NLTK's tokenizer ~\cite{Bird:2009:NLP:1717171},
while for Arabic we used Farasa's segmenter~\cite{abdelali2016farasa}. For Arabic, tokenization is not enough; we also need word segmentation since conjunctions and clitics are commonly attached to the main word, e.g., \AR{ـه ُ} + \AR{بَيتـُ} + \AR{وَ} 
(`and his house', lit. ``and house his''). This causes explosion in the vocabulary size and data sparseness.

\noindent We further needed a part-of-speech (POS) tagger for Arabic, for which we used Farasa \cite{abdelali2016farasa}, while we used NLTK's POS tagger for English ~\cite{Bird:2009:NLP:1717171}. 
This yields different tagsets: for English, this is the Penn Treebank tagset~\cite{marcus1993building}, while for Arabic this the Farasa tagset. Thus, we had to further map all POS tags to the same tagset: the Universal tagset~\cite{petrov2011universal}.

\subsection{Evaluation}




We train the system on five English political debates, and we test on two debates: either English or their Arabic translations. Note that, compared to our original model~\cite{gencheva-EtAl:2017:RANLP}, here we use more debates: seven instead of four. Moreover, here we exclude some of the features, namely some debate-specific information (e.g., speaker, system messages), in order to be able to process any free text, and also discourse parse features, as we do not have a discourse parser for Arabic.

One of the most important components of the system that we had to port across languages were the word embeddings. We experimented with the following cross-language embeddings:

\noindent -- \emph{VecMap}: we used a parallel English-Arabic corpus of TED talks\footnote{We used TED talks as they are conversational large corpora, which is somewhat close to the debates we train on.} \cite{cettoloEtAl:EAMT2012} to generate monolingual embeddings (Arabic and English) using word2vec~\cite{mikolov2013distributed}. Then we projected these embeddings into a joint vector space using VecMap~\cite{artetxe2017learning}.

\noindent -- \emph{MUSE embeddings}: In a similar fashion, we generated cross-language embeddings from the same TED talks using Facebook's supervised MUSE model~\cite{lample2017unsupervised} to project the Arabic and the English monolingual embeddings into a joint vector space.

\noindent -- \emph{Attract-Repel embeddings}: we used the pre-trained English-Arabic embeddings from Attract-Repel~\cite{Mrksic:2017}.


Table~\ref{table:combination} shows the system performance when predicting claims by any of the sources, using word2vec and the cross-language embeddings.\footnote{Note that these results are not comparable to those in~\cite{gencheva-EtAl:2017:RANLP} as we use a different evaluation setup: train/test split vs. cross-validation, debates that involve not only Hillary Clinton and Donald Trump, and we also disable the metadata and the discourse parse features.} All results are well above a random baseline.

\begin{figure*}	
\footnotesize
\centering
\includegraphics[width=0.93 \textwidth,keepaspectratio]{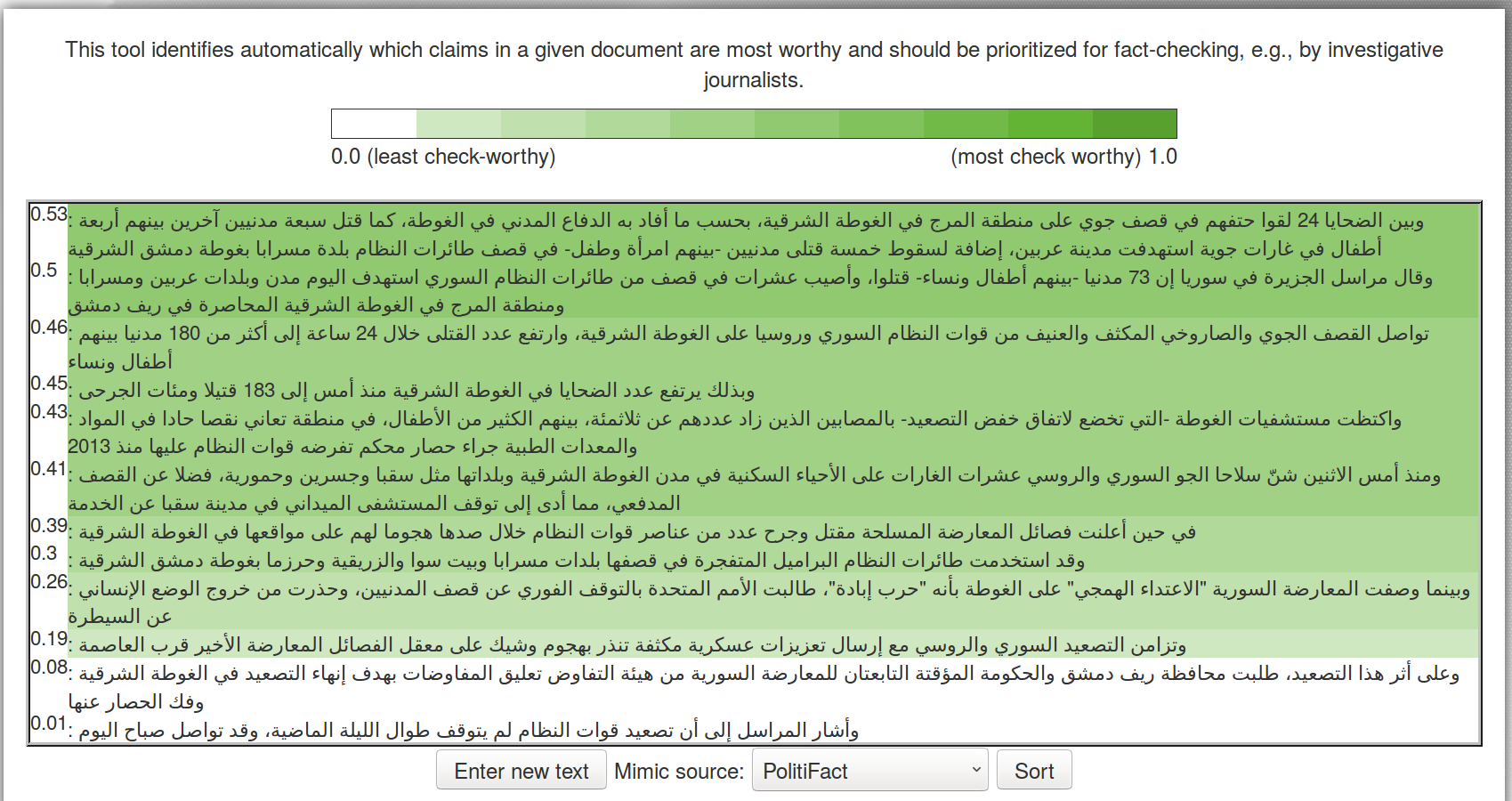}
\caption{\label{fig:demoArabic} Screenshot of ClaimRank's output for an Arabic news article,
sorted by score.}
\end{figure*}

\begin{table*}
\small
\centering
\begin{tabular}{l|c@{\hspace{3mm}} c@{\hspace{3mm}}c@{\hspace{3mm}}c@{\hspace{3mm}}c@{\hspace{3mm}}c@{\hspace{3mm}}|	c@{\hspace{3mm}}c@{\hspace{3mm}}c@{\hspace{3mm}}c@{\hspace{3mm}}c@{\hspace{3mm}}c@{\hspace{3mm}}}
    & \multicolumn{6}{c}{\bf \emph{English}} & \multicolumn{6}{c}{\bf \emph{Arabic}}\\
    \bf System & \bf MAP & \bf R-Pr & \bf P@5 & \bf P@10 & \bf P@20 & \bf P@50 & \bf MAP & \bf R-Pr & \bf P@5 & \bf P@10 & \bf P@20 & \bf P@50 \\ \hline
     word2vec      &0.323&0.330&0.80&0.60&0.45&0.38& --- & --- & --- & --- & --- & ---\\ 
     \hline
     VecMap      &0.298&0.333&0.30&0.40&0.45&0.44&0.291&0.324&0.10&0.25&0.35&0.41\\ 
     MUSE      &0.319&0.332&0.40&0.45&0.50&0.49&0.302&0.331&0.10&0.25&0.38&0.48\\ 
     Attract-Repel      &0.342&0.385&0.40&0.45&0.50&0.46&0.263&0.312&0.10&0.15&0.30&0.41\\ 
    \hline
    Random&0.161&0.161&0.10&0.20&0.13&0.08\\
  \end{tabular}
\caption{Performance when using different cross-language embeddings.}
\label{table:combination}
\end{table*}

We can see some drop in MAP for English when using VecMap or MUSE, which is to be expected as the model needs to balance between preserving the original embeddings and projecting them into a joint space. The Attract-Repel vectors perform better for English, which is probably due to the monolingual synonymy/antonymy constraints that they impose~\cite{P17-1006}, thus yielding better vectors, even for English.

The overall MAP results for Arabic are competitive, compared to English.
The best model 
is MUSE, while Attract-Repel is way behind, probably because, unlike VecMap and MUSE, its word embeddings are trained on unsegmented Arabic, which causes severe data sparseness issues.

\noindent In the final system, we use MUSE vectors for both languages, which perform best overall: not only for MAP, but also P@20, and P@50, which are very important measures assuming that manual fact-checking can be done for up to 20 or up to 50 claims only (in fact, statistics show that eight out of our nine fact-checking organizations had no more than 50 claims checked per debate).

\section{The System in Action}

ClaimRank is available online.\footnote{\url{ http://claimrank.qcri.org}}
Our systems' user interface 
consists of three views:

\noindent \emph{-- The text entry view}: composed of a text box, and a submit button. 

\noindent \emph{-- The results view} shows the text split into sentences with scores reflecting the degree of check-worthiness, and each sentence has a color intensity that reflects its score range, as shown in Figure~\ref{fig:demoEnglish}. The user can sort the results, or choose to mimic different media.


\noindent \emph{-- The sorted results view} shows the most check-worthy sentences first, as Figure~\ref{fig:demoArabic} shows.

\section{Conclusion and Future Work}

We have presented ClaimRank ---an online system for prioritizing check-worthy claims. ClaimRank can help professional fact-checkers and journalists in their work as it can help them identify where they should focus their efforts first. The system learns from selections by nine reputable fact-checking organizations, and as a result, it can mimic the sentence selection strategies as applied by each and any of them, as well as for the union of them all.

While originally trained on a collection of political debates, ClaimRank can also work for other kinds of text, e.g.,~interviews or just regular news articles. 
Moreover, even though initially developed for English, the system was subsequently adapted to also support Arabic, using a combination of manual training data translation and cross-language embeddings. 


In future work, we wold like to train the models on more political debates and speeches, as well as on other genres. We further plan to add support for more languages. 



\bibliography{naaclhlt2018}

\begin{thebibliography}{21}
\expandafter\ifx\csname natexlab\endcsname\relax\def\natexlab#1{#1}\fi

\bibitem[{Abdelali et~al.(2016)Abdelali, Darwish, Durrani, and
  Mubarak}]{abdelali2016farasa}
Ahmed Abdelali, Kareem Darwish, Nadir Durrani, and Hamdy Mubarak. 2016.
\newblock Farasa: A fast and furious segmenter for {A}rabic.
\newblock In \emph{Proceedings of the Conference of the North American Chapter
  of the Association for Computational Linguistics}, NAACL-HLT~'16, pages
  11--16, San Diego, CA, USA.

\bibitem[{Artetxe et~al.(2017)Artetxe, Labaka, and
  Agirre}]{artetxe2017learning}
Mikel Artetxe, Gorka Labaka, and Eneko Agirre. 2017.
\newblock Learning bilingual word embeddings with (almost) no bilingual data.
\newblock In \emph{Proceedings of the 55th Annual Meeting of the Association
  for Computational Linguistics}, ACL~'17, pages 451--462, Vancouver, Canada.

\bibitem[{Bird et~al.(2009)Bird, Klein, and Loper}]{Bird:2009:NLP:1717171}
Steven Bird, Ewan Klein, and Edward Loper. 2009.
\newblock \emph{Natural Language Processing with Python}, 1st edition.
\newblock O'Reilly Media, Inc.

\bibitem[{Blei et~al.(2003)Blei, Ng, and Jordan}]{blei2003latent}
David~M Blei, Andrew~Y Ng, and Michael~I Jordan. 2003.
\newblock Latent {D}irichlet allocation.
\newblock \emph{Journal of Machine Learning Research}, 3(Jan):993--1022.

\bibitem[{Cettolo et~al.(2012)Cettolo, Girardi, and
  Federico}]{cettoloEtAl:EAMT2012}
Mauro Cettolo, Christian Girardi, and Marcello Federico. 2012.
\newblock {WIT}$^3$: Web inventory of transcribed and translated talks.
\newblock In \emph{Proceedings of the 16th Conference of the European
  Association for Machine Translation}, EAMT~'12, pages 261--268, Trento,
  Italy.

\bibitem[{Gencheva et~al.(2017)Gencheva, Nakov, M\`{a}rquez,
  Barr\'{o}n-Cede\~{n}o, and Koychev}]{gencheva-EtAl:2017:RANLP}
Pepa Gencheva, Preslav Nakov, Llu\'{i}s M\`{a}rquez, Alberto
  Barr\'{o}n-Cede\~{n}o, and Ivan Koychev. 2017.
\newblock A context-aware approach for detecting worth-checking claims in
  political debates.
\newblock In \emph{Proceedings of the International Conference Recent Advances
  in Natural Language Processing}, RANLP~'17, pages 267--276, Varna, Bulgaria.

\bibitem[{Glorot et~al.(2011)Glorot, Bordes, and Bengio}]{pmlr-v15-glorot11a}
Xavier Glorot, Antoine Bordes, and Yoshua Bengio. 2011.
\newblock Deep sparse rectifier neural networks.
\newblock In \emph{Proceedings of the Fourteenth International Conference on
  Artificial Intelligence and Statistics}, PMLR~'15, pages 315--323, Fort
  Lauderdale, FL, USA.

\bibitem[{Hassan et~al.(2015)Hassan, Li, and Tremayne}]{Hassan:15}
Naeemul Hassan, Chengkai Li, and Mark Tremayne. 2015.
\newblock Detecting check-worthy factual claims in presidential debates.
\newblock In \emph{Proceedings of the 24th ACM International Conference on
  Information and Knowledge Management}, CIKM~'15, pages 1835--1838, Melbourne,
  Australia.

\bibitem[{Hooper(1974)}]{hooper1974assertive}
Joan~B. Hooper. 1974.
\newblock \emph{On Assertive Predicates}.
\newblock Indiana University Linguistics Club. Indiana University Linguistics
  Club.

\bibitem[{Joty et~al.(2015)Joty, Carenini, and Ng}]{jotycodra}
Shafiq Joty, Giuseppe Carenini, and Raymond~T. Ng. 2015.
\newblock {CODRA}: A novel discriminative framework for rhetorical analysis.
\newblock \emph{Comput. Linguist.}, 41(3):385--435.

\bibitem[{Lample et~al.(2017)Lample, Denoyer, and
  Ranzato}]{lample2017unsupervised}
Guillaume Lample, Ludovic Denoyer, and Marc'Aurelio Ranzato. 2017.
\newblock Unsupervised machine translation using monolingual corpora only.
\newblock \emph{arXiv preprint arXiv:1711.00043}.

\bibitem[{LeCun et~al.(1998)LeCun, Bottou, Bengio, and
  Haffner}]{lecun1998gradient}
Yann LeCun, L{\'e}on Bottou, Yoshua Bengio, and Patrick Haffner. 1998.
\newblock Gradient-based learning applied to document recognition.
\newblock \emph{Proceedings of the IEEE}, 86(11):2278--2324.

\bibitem[{Liu et~al.(2005)Liu, Hu, and Cheng}]{Liu:2005:OOA:1060745.1060797}
Bing Liu, Minqing Hu, and Junsheng Cheng. 2005.
\newblock Opinion observer: Analyzing and comparing opinions on the web.
\newblock In \emph{Proceedings of the 14th International Conference on World
  Wide Web}, WWW '05, pages 342--351, New York, NY, USA.

\bibitem[{Loper and Bird(2002)}]{bird2004nltk}
Edward Loper and Steven Bird. 2002.
\newblock {NLTK}: The natural language toolkit.
\newblock In \emph{Proceedings of the Workshop on Effective Tools and
  Methodologies for Teaching Natural Language Processing and Computational
  Linguistic}, ETMTNLP~'02, pages 63--70, Philadelphia, PA, USA.

\bibitem[{Marcus et~al.(1993)Marcus, Marcinkiewicz, and
  Santorini}]{marcus1993building}
Mitchell~P Marcus, Mary~Ann Marcinkiewicz, and Beatrice Santorini. 1993.
\newblock Building a large annotated corpus of {E}nglish: The {P}enn
  {T}reebank.
\newblock \emph{Computational linguistics}, 19(2):313--330.

\bibitem[{Mikolov et~al.(2013)Mikolov, Sutskever, Chen, Corrado, and
  Dean}]{mikolov2013distributed}
Tomas Mikolov, Ilya Sutskever, Kai Chen, Greg~S Corrado, and Jeff Dean. 2013.
\newblock Distributed representations of words and phrases and their
  compositionality.
\newblock In \emph{Advances in neural information processing systems},
  NIPS~'13, pages 3111--3119, Lake Tahoe, CA, USA.

\bibitem[{Mrk\v{s}i\'c et~al.(2017)Mrk\v{s}i\'c, Vuli\'{c}, {\'O S\'eaghdha},
  Leviant, Reichart, Ga\v{s}i\'c, Korhonen, and Young}]{Mrksic:2017}
Nikola Mrk\v{s}i\'c, Ivan Vuli\'{c}, Diarmuid {\'O S\'eaghdha}, Ira Leviant,
  Roi Reichart, Milica Ga\v{s}i\'c, Anna Korhonen, and Steve Young. 2017.
\newblock Semantic specialisation of distributional word vector spaces using
  monolingual and cross-lingual constraints.
\newblock \emph{Transactions of the Association for Computational Linguistics},
  5:309--324.

\bibitem[{Patwari et~al.(2017)Patwari, Goldwasser, and Bagchi}]{Patwari:17}
Ayush Patwari, Dan Goldwasser, and Saurabh Bagchi. 2017.
\newblock {TATHYA:} {A} multi-classifier system for detecting check-worthy
  statements in political debates.
\newblock In \emph{Proceedings of the {ACM} on Conference on Information and
  Knowledge Management}, CIKM~'17, pages 2259--2262, Singapore.

\bibitem[{Petrov et~al.(2012)Petrov, Das, and McDonald}]{petrov2011universal}
Slav Petrov, Dipanjan Das, and Ryan McDonald. 2012.
\newblock A universal part-of-speech tagset.
\newblock In \emph{Proceedings of the Eight International Conference on
  Language Resources and Evaluation}, LREC~'12, pages 2089--2096, Istanbul,
  Turkey.

\bibitem[{Recasens et~al.(2013)Recasens, Danescu-Niculescu-Mizil, and
  Jurafsky}]{Recasens+al:13a}
Marta Recasens, Cristian Danescu-Niculescu-Mizil, and Dan Jurafsky. 2013.
\newblock Linguistic models for analyzing and detecting biased language.
\newblock In \emph{Proceedings of the 51st Annual Meeting of the Association
  for Computational Linguistics}, ACL~'13, pages 1650--1659, Sofia, Bulgaria.

\bibitem[{Vuli{\'{c}} et~al.(2017)Vuli{\'{c}}, Mrk{\v{s}}i{\'{c}}, Reichart,
  {\'O}~S{\'e}aghdha, Young, and Korhonen}]{P17-1006}
Ivan Vuli{\'{c}}, Nikola Mrk{\v{s}}i{\'{c}}, Roi Reichart, Diarmuid
  {\'O}~S{\'e}aghdha, Steve Young, and Anna Korhonen. 2017.
\newblock Morph-fitting: Fine-tuning word vector spaces with simple
  language-specific rules.
\newblock In \emph{Proceedings of the 55th Annual Meeting of the Association
  for Computational Linguistics}, ACL~'17, pages 56--68, Vancouver, Canada.

\end{thebibliography}
\bibliographystyle{acl_natbib}

\end{document}